\documentclass[journal]{IEEEtran}
\ifCLASSINFOpdf
   \usepackage[pdftex]{graphicx}
\else
\fi
\usepackage{amsmath}
\usepackage[pagebackref=false,breaklinks=true,hidelinks,bookmarks=false]{hyperref}

\usepackage{url}

\usepackage{multirow}

\begin{document}
%
\title{L1-regularized Reconstruction \\ Error as Alpha Matte}
%
%
%

\author{Jubin~Johnson,
        Hisham~Cholakkal,
        and~Deepu~Rajan
\thanks{J. Johnson, H. Cholakkal, and D. Rajan are with the Multimedia Lab, School of Computer Science and
Engineering, Nanyang Technological University, Singapore,
639798 (e-mail: \{jubin001, hisham002, asdrajan\}@ntu.edu.sg).}
}

\maketitle

\begin{abstract}
Sampling-based alpha matting methods have traditionally followed the compositing equation to estimate the alpha value at a pixel from a pair of foreground (F) and background (B) samples. The (F,B) pair that produces the least reconstruction error is selected, followed by alpha estimation. The significance of that residual error has been left unexamined. In this letter, we propose a video matting algorithm that uses L1-regularized reconstruction error of F and B samples as a measure of the alpha matte. A multi-frame non-local means framework using coherency sensitive hashing is utilized to ensure temporal coherency in the video mattes. Qualitative and quantitative evaluations on a dataset exclusively for video matting demonstrate the effectiveness of the proposed matting algorithm.          
\end{abstract}


%
\IEEEpeerreviewmaketitle

\section{Introduction}
%
%
%
%
\IEEEPARstart{D}{igital} matting refers to the problem of accurate foreground extraction and finds its use in image and video editing. Mathematically, any pixel color $I_i$ can be modeled as a convex combination of the foreground color ($F_i$) and the background color ($B_i$) such that 
\begin{equation}
I_i = \alpha_iF_i+(1-\alpha_i)B_i,
\label{eq:matting}
\end{equation}
where $\alpha_i$ is the opacity (alpha) value at pixel $i$. Determining $\alpha$ is an under-constrained problem, made tractable by means of user-input labels in the form of a trimap or scribbles.

Matting methods are generally divided into sampling-based~\cite{wang2007optimized,he2011global,shahrian2013improving,johnson2014sparse} and propagation-based~\cite{levin2008closed,chen2012knn} approaches. The former category uses color values from the known foreground and background regions to find the best foreground-background $(F,B)$ pair that represents the true foreground and background colors to estimate $\alpha$ of a given pixel. Different sampling strategies (local/global) and optimization criteria for selecting the best $(F,B)$ pair distinguish these approaches. Similar color distribution among the foreground and background regions is a challenge since the samples cannot discriminate between $F$ and $B$ regions anymore. Propagation-based methods leverage the correlation between neighboring pixels with respect to local image statistics to interpolate the known alpha values to the unknown regions. As with sampling approaches, false correlations between neighboring $F$ and $B$ pixels occurs due to color similarity. Moreover, strong edges and textured regions fail to propagate the alpha accurately. Recently, deep learning based approaches~\cite{cho2016natural} have shown to perform well in natural image matting.

\begin{figure}[t]
\centering
\includegraphics[width=1\linewidth, clip=true, trim=0.1cm 1.6cm 14.3cm 2cm]{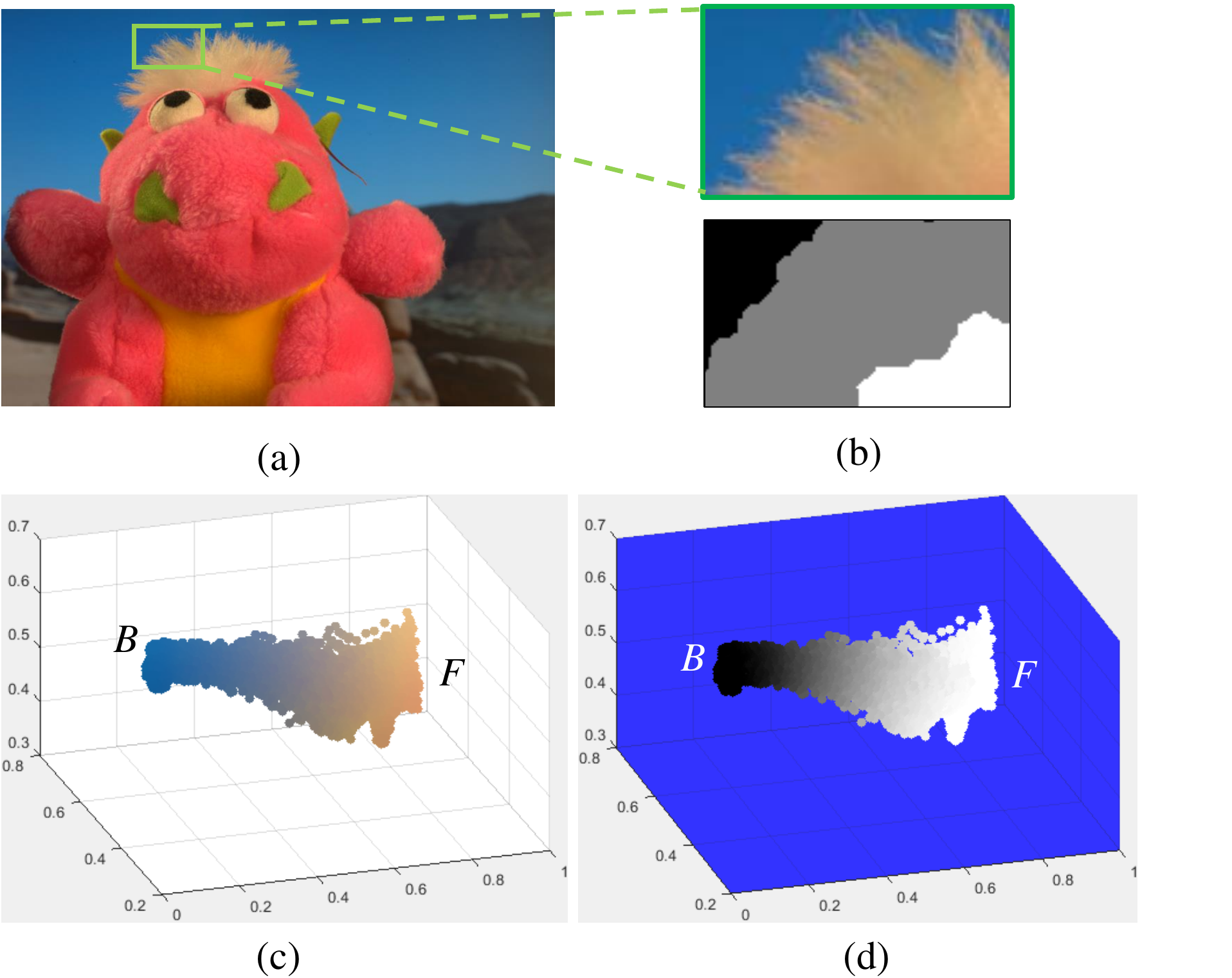}
\caption{Illustration of reconstruction error as alpha matte. (a) Input image with a (b) zoomed window and its trimap, (c) color distribution of the pixels in RGB color space indicating definite and unknown regions, and (d) plot of alpha generated from reconstruction error of pixels with values in $[0,1]$.}
\label{fig:overview}
\end{figure}
Video matting, apart from extracting spatially accurate mattes on each frame, also has the additional requirement of temporal coherence across the video~\cite{sharianvideo,bai2011towards,johnsontip}. The human visual system is highly sensitive to jitter and temporal inconsistencies across frames. Low contrast and fast motion are factors that contribute to inaccurate matte in a frame, thereby leading to temporal jitter across the extracted video matte. 
 Although the quality of the mattes obtained by independently applying image matting algorithms to each frame is high, it does not result in temporally coherent mattes. $\alpha$-propagation has been extended to the temporal domain as post-processing to alleviate this problem. Snapcut~\cite{bai2009video} uses the matting Laplacian~\cite{levin2008closed} to bias the alpha to the previous frame. A motion-aware Laplacian is constructed to propagate the matte temporally in \cite{li2013motion}. Level-set interpolation is used to temporally smooth the estimated mattes in \cite{bai2011towards}.  Optical flow is used to warp the alpha from the previous frame in the Laplacian formulation in \cite{sharianvideo}.       

\begin{figure*}[t]
{\centering
\includegraphics[width=1\linewidth, clip=true, trim=0.1cm 7.6cm 0.1cm 5.65cm]{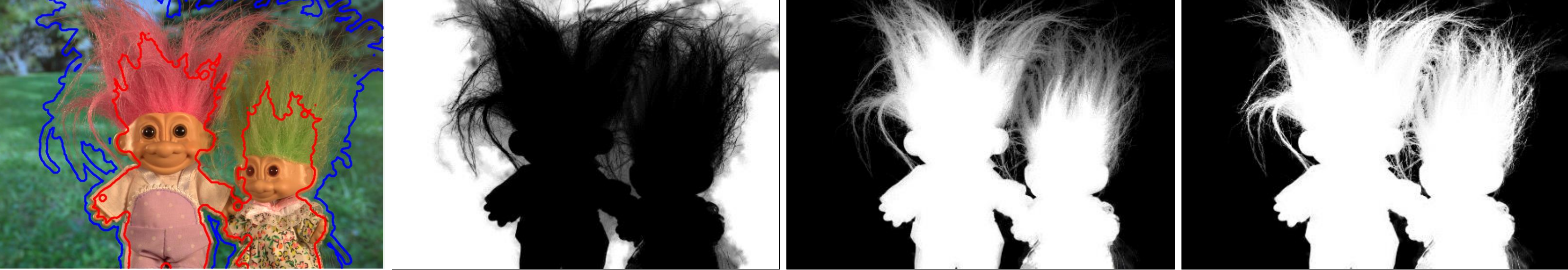}}
\hspace*{2cm}(a)\hspace*{4cm} (b) \hspace*{4cm}(c) \hspace*{4cm} (d)
\caption{L1-regularized reconstruction error as alpha matte. (a) Input image and its trimap, plots of (b) $\xi_F$, (c) $\xi_B$ and (d) estimated alpha map using eq.~(\ref{eq:alphaest}).}
\label{fig:pbmap}
\end{figure*}
The proposed approach is based on sampling. As mentioned earlier, sampling methods find the best $(F,B)$ pair that satisfies eq.~(\ref{eq:matting}) and use it to estimate the alpha value. The reconstruction error in the selected pair is $\xi_i=\left \| I_i-(\hat{\alpha}_iF_i+(1-\hat{\alpha}_i)B_i) \right \|$. The significance of this residual error for matting has largely been left unexamined in literature.  Johnson~\textit{et~al}.~\cite{johnsontip} showed sparse coding as an alternative to the compositing equation for estimating the $\alpha$ value at a pixel. Inspired by this, we propose a sampling-based approach that looks at matting from the perspective of sparse reconstruction error of feature samples. Fig.~\ref{fig:overview} illustrates the motivation behind using reconstruction error as a measure of the matte in a real image. A zoomed region of the input image in Fig.~\ref{fig:overview}(a) and its trimap are shown in Fig.~\ref{fig:overview}(b), representing a hairy region containing mixed pixels. The local smoothness assumption between the alpha values of neighboring pixels is paramount to extracting a good matte. In a real image, alpha would gradually transition between the definite $F$ and $B$ with the true mixed pixel alphas having an intermediate value. The RGB color distribution of pixels in the image patch varies smoothly between the foreground and background with the blending peaking at the middle of the unknown region (Fig.~\ref{fig:overview}(c)). Similarly, the error obtained during reconstruction using $F$ and $B$ samples can be thought of as a probability measure that varies smoothly between the foreground and background regions, gradually rising from the definite regions and peaking at true mixed pixels. As can be seen in Fig.~\ref{fig:overview}(c) and (d), the color distribution of pixels in a real image and the residual error are highly correlated. To the best of our knowledge, we are the first to formulate matting from the perspective of reconstruction error.  A patch-based non-local means (NLM) framework using coherency sensitive hashing across multiple frames is integrated into the estimated mattes to ensure temporal coherence in the final mattes. The proposed NLM framework is shown to reduce temporal jitter when compared to the widely used Laplacian methods using qualitative and quantitative comparisons on a video matting dataset~\cite{sharianvideo,johnsontip}.                                          
\vspace*{-0.1cm}
\section{Proposed Approach}
\subsection{L1-regularized reconstruction error as alpha matte}
The aim of the proposed method is to use reconstruction error as a measure of the $\alpha$ value. The objective of using error reconstruction hinges on the assumption that the foreground and background are locally smooth, akin to propagation-based methods. The idea is therefore, to select a local subset of the known regions for the local smoothness assumption to hold. Following \cite{johnsontip}, at each pixel, $F$ and $B$ dictionaries are formed by sampling the spatially nearest pixels at a radius of 50 pixels from the definite foreground and background regions, respectively. The feature vector used is the 8-D vector $[\:R\; G \;B \;L\; a \;b \; x \;y \:]^{T}$ formed by concatenating the RGB and CIELAB color-spaces along with the $X$-$Y$ coordinates. In order to reduce the sample space, the definite $F$ and $B$ regions are clustered into superpixels using SLIC segmentation~\cite{achanta2012slic}. It is to be noted that \cite{johnsontip} uses a single dictionary by concatenating the $F$ and $B$ samples together. However, the proposed method requires separate $F$ and $B$ dictionaries in order to determine the reconstruction error with respect to each as explained below. 

Given an unknown pixel $i$, let $D_F^i$ and $D_B^i$ be the foreground and background dictionaries formed by sampling the feature vectors. The sparse codes with respect to each dictionary are determined as 
\begin{equation}
\beta_F^i=argmin\left \| v_{i}-D_F^i\beta_F^{i} \right \|_{2}^{2} + \lambda\left \| \beta_F^{i} \right \|_{1},
\end{equation} 
\begin{equation}
\beta_B^i=argmin\left \| v_{i}-D_B^i\beta_B^{i} \right \|_{2}^{2} + \lambda\left \| \beta_B^{i} \right \|_{1},
\end{equation}     
where $v_i$ is the feature vector at pixel $i$. The residual errors generated by reconstruction using $F$ and $B$ dictionaries are
\begin{equation}
\xi_F^i = \left \| v_{i}-D_F^i\beta_F^{i} \right \|_{2}, \;\; \xi_B^i = \left \| v_{i}-D_B^i\beta_B^{i} \right \|_{2}.
\end{equation}
$\xi_F^i$ $(\xi_B^i)$ is the error generated at the unknown pixel $i$ when its feature is reconstructed using foreground (background) dictionary. A high value for $\xi_F^i$ $(\xi_B^i)$ indicates that the current pixel cannot be reconstructed well enough by the $F$ $(B)$ samples. Fig.~\ref{fig:pbmap} (b) and (c) visualizes these error maps for a real image. $\xi_F^i$ should ideally be 0 for foreground pixels and gradually increase towards the background pixels. Similarly, $\xi_B^i$ should ideally be 0 for background pixels and gradually increase towards foreground regions. A pixel with a true alpha value of 0.5, i.e. a truly mixed pixel should have comparable reconstruction errors in $\xi_F^i$ and $\xi_B^i$.     

The alpha value can be interpreted as the probability of the pixel belonging to the foreground. $\xi_B^i$ represents the probability of belonging to the foreground, given the known background information, i.e., $\xi_B^i = P(f(i)|\mathbf{D}_B)$. $\xi_F^i$ represents the probability of belonging to the background, given the known foreground information - $\xi_F^i = P(b(i)|\mathbf{D}_F) = 1-P(f(i)|\mathbf{D}_F)$.
Based on the above observation, the alpha value is then estimated as   
\begin{equation}
\hat{\alpha}_i= \frac{P(f(i)|\mathbf{D}_B)}{P(f(i)|\mathbf{D}_B)+P(b(i)|\mathbf{D}_F)}=\frac{\xi_B^i}{\xi_B^i+\xi_F^i}.
\label{eq:alphaest}
\end{equation}  
\begin{figure}[t]
\centering
\includegraphics[width=1\linewidth, clip=true, trim=0.2cm 2.1cm 0.1cm 3.7cm]{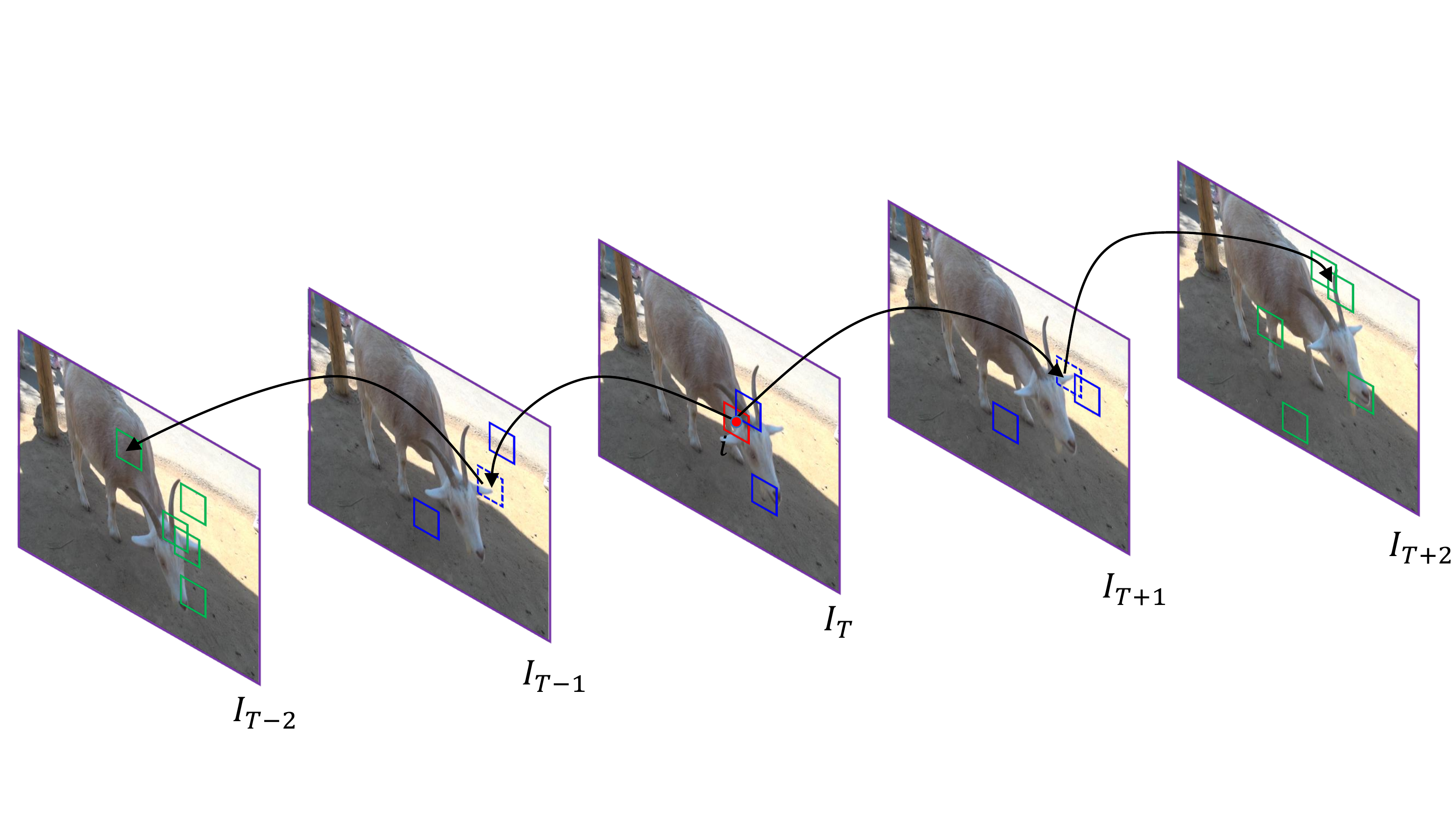}
\caption{Illustration of our NLM framework to maintain temporal coherence across the video. For the red patch, AKNN's from immediate frames (blue) are used to find coherent patches from the next frames (green).}
\label{fig:NLMoverview}
\end{figure}
\begin{figure*}[t]
{\centering
\includegraphics[width=1\linewidth, clip=true, trim=0.5cm 11.1cm 2.3cm 0.3cm]{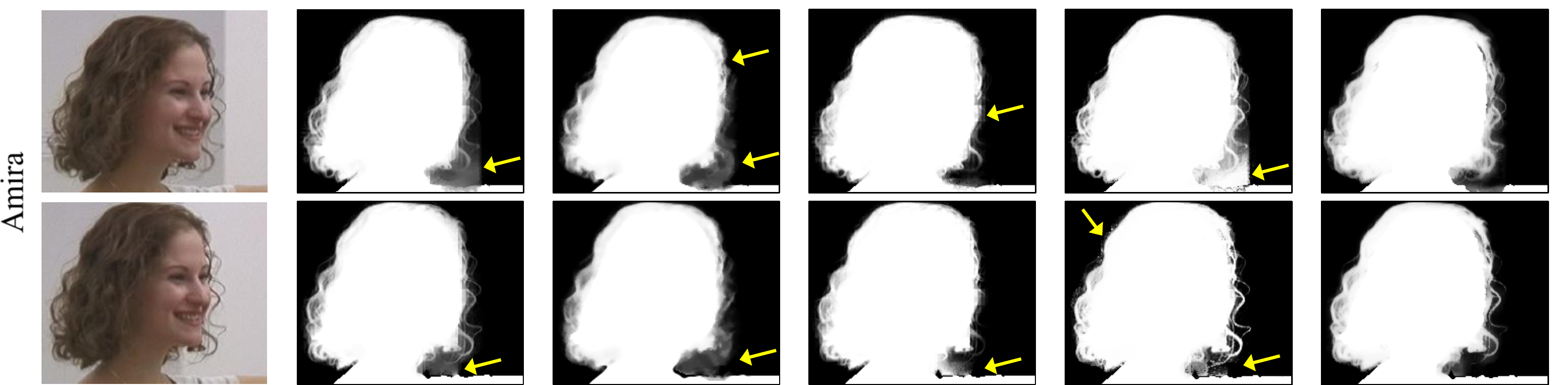}}
\hspace*{1.6cm}(a)\hspace*{2.45cm} (b) \hspace*{2.45cm}(c) \hspace*{2.4cm} (d) \hspace*{2.3cm} (e) \hspace*{2.3cm} (f)
\caption{Qualitative comparison of the video mattes on \emph{Amira} sequence. (a) Input frames, mattes of (b) SC~\cite{bai2009video}, (c) BA~\cite{bai2011towards}, (d) EH~\cite{sharianvideo}, (e) JO~\cite{johnsontip} and (f) proposed method. Yellow arrows indicate discontinuities in the matte.}
\label{fig:Qualitative}
\end{figure*} 
As can be seen from eq.~(\ref{eq:alphaest}), if a pixel truly belongs to the foreground, its foreground reconstruction error $\xi_F^i$ will be a smaller value than the background reconstruction error $\xi_B^i$, thereby ensuring $\alpha$ is large.                              
The alpha map is shown in Fig.~\ref{fig:pbmap}(d) and indicates the effectiveness of this simple formulation using sparse reconstruction error.    

\subsection{Patch-based non-local means for temporal coherence}
Since the sampling strategy uses a local spatial subset of samples from within the frame, the alpha estimates obtained above lack temporal coherency as the information present in the nearby frames is ignored. Existing methods follow a Laplacian based post-processing step where the inter-pixel correlation is utilized to propagate the matte. The disadvantage inherently lies in its inability to find distant neighbors in space and time. Also, the use of pixel-based matching leads to noise from outliers that get matched incorrectly. To handle this, we propose a patch-based NLM framework that is prevalent in video denoising~\cite{liu2010high} to maintain the temporal consistency across neighboring frames. NLM~\cite{buades2005non} was originally introduced to remove noise by averaging pixels in an image weighted by local patch similarities. The high search complexity in finding non-local neighboring patches restricts its use to a local neighborhood alone. Therefore, we apply an approximate K-nearest neighbor patch-matching using coherency sensitive hashing~\cite{korman2011coherency} that extends PatchMatch~\cite{barnes2010generalized} using a hashing scheme where similar patches in the temporal neighborhood are used to propagate the matches to their neighbors. 

The framework is illustrated in Fig.~\ref{fig:NLMoverview}. For a given image patch $P_T(i)$ (shown in red) centered at pixel $i$ in frame $T$, approximate K-nearest neighbors (AKNN) in frames $T-1$ and $T+1$ (in blue) are initialized by creating $L$ hash tables based on projection of the patches on Walsh-Hadamard kernels, followed by search for the best candidate patches~\cite{korman2011coherency}. The two images are assumed to be coherent, i.e., for every pair of similar patches, their neighbors are also likely to be similar. In the next iteration, the AKNN field for the initial candidates are searched for in their respective temporal neighbors $T-2$ and $T+2$. Under this setup, the temporal neighbors for $P_t(z)$ include a series of AKNN's $\{\mathcal{N}_{T-2},\mathcal{N}_{T-1},\mathcal{N}_{T},\mathcal{N}_{T+1},\mathcal{N}_{T+2}\}$. $\mathcal{N}_i=\{P(z_{tj})\}_{j=1}^{K}$ denotes the patches in the $t^{th}$ frame. The non-local means estimate of alpha is~\cite{liu2010high} 
\begin{equation}
\alpha_T(i)=\frac{1}{\Omega}\sum_{t=T-2}^{T+2}\gamma^{\left | t-T \right |}\sum_{j=1}^{K}\hat{\alpha}(i_{tj})\textup{exp}\left \{ -\frac{D_w(P(i),P(i_{tj}))}{2\sigma_t^2} \right \},
\label{eq:NLM}
\end{equation}
where $\hat{\alpha}(i)$ is the alpha value of a patch centered at $i$ and $\Omega$ is a normalization constant:
\begin{equation}
\Omega=\sum_{t=T-2}^{T+2}\gamma^{\left | t-T \right |}\sum_{j=1}^{K}\textup{exp}\left \{ -\frac{D_w(P(i),P(i_{tj}))}{2\sigma_t^2} \right \}.
\end{equation}           
$D_w(.,.)$ is a weighted sum of squared difference (SSD) over 2 patches denoted by 	 
\begin{equation}
\begin{aligned}
D_w\left (P(i_1),P(i_2)  \right ) = \qquad \qquad \qquad\qquad\qquad \\
\sum_{u\in[-s,s]^2}\left (P(i_1+u)-P(i_2+u  \right ))^2
\textup{exp}\left \{ -\frac{\left \| u \right \|^2}{2\sigma_p^2} \right \},
\end{aligned}
\end{equation} 
where $s$ is the patch width and $\sigma_p$ is set to $s/2$. $\gamma$ is set to 0.9 to control the influence of temporal neighbors. Eq.~(\ref{eq:NLM}) estimates the alpha matte for an entire patch centered at pixel $i$ in frame $T$. For an alpha patch of width $s$, all patches whose centers are located within a radius of $\frac{s}{2}$ from $i$, contain the pixel at $i$. A simple Gaussian weighted averaging is performed to obtain the final NLM estimate at $i$ as $\alpha_i=\frac{{}\sum_{j\in\phi }\alpha(i)e^{-\frac{\left \| i-i_j \right \|^2}{2\sigma_p^2}}}{\sum_{j\in\phi }e^{-\frac{\left \| i-i_j \right \|^2}{2\sigma_p^2}}}$
where $\phi$ denotes the set of patches that contain the pixel $i$. The process is repeated for each frame of the video to obtain the final video mattes. 
\begin{figure}[t]
{\centering
\includegraphics[width=1\linewidth, clip=true, trim=0.2cm 6.5cm 0.1cm 2.5cm]{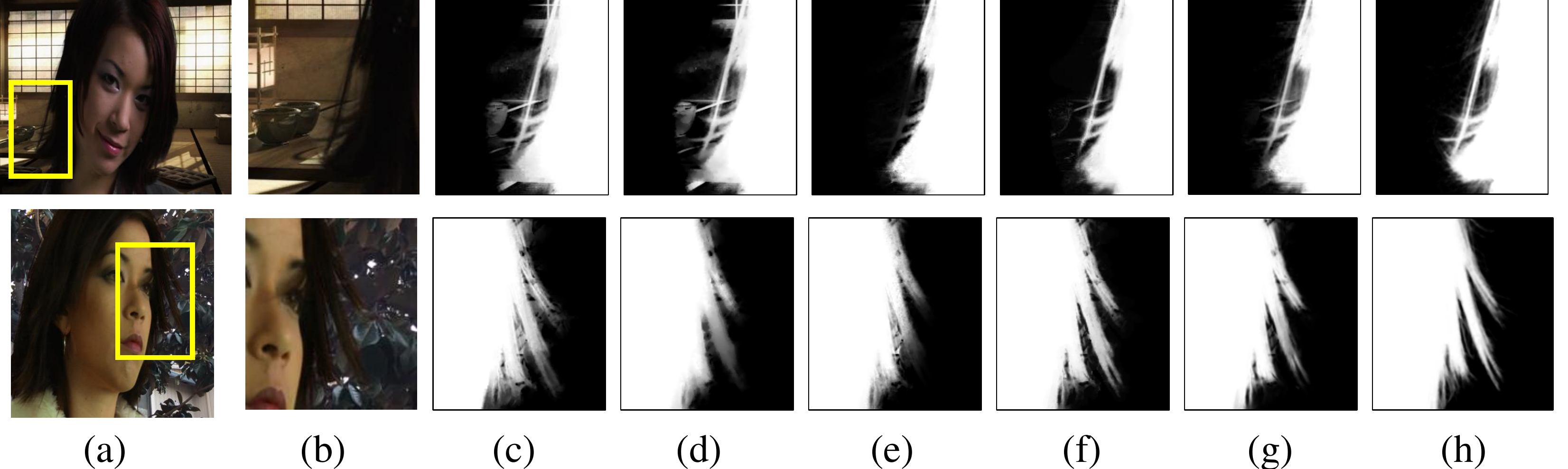}}
\caption{Qualitative comparison of the video mattes with ground truth on \emph{Face} and \emph{Woman} sequences. (a) Input frames and (b) zoomed region, mattes of (c) SC~\cite{bai2009video}, (d) BA~\cite{bai2011towards}, (e) EH~\cite{sharianvideo}, (f) JO~\cite{johnsontip} , (g) proposed method and (h) Ground truth.}
\label{fig:Qualitative2}
\end{figure}
\section{Experimental Results} 
The effectiveness of the proposed method is evaluated on an exclusive video matting dataset used in \cite{sharianvideo,johnsontip}. It contains sequences covering a wide range of pixel opacity variations and challenges like occlusion and low-contrast. Trimaps are generated on each frame using the method of \cite{johnson2015temporal}. $\lambda$ for sparse coding is set to 0.1. In all experiments, the patch size was set to $8\times8$. K is set to 5 in eq.~(\ref{eq:NLM}). The proposed method is evaluated both quantitatively and qualitatively with recent video matting approaches namely, Snapcut (SC)~\cite{bai2009video}, Bai \emph{et al}. (BA)~\cite{bai2011towards}, Ehsan \emph{et al}. (EH)~\cite{sharianvideo} and Johnson \emph{et al}. (JO)~\cite{johnsontip}. 
\subsection{Qualitative comparison}
Fig.~\ref{fig:Qualitative} and Fig.~\ref{fig:Qualitative2} show the visual comparison of the proposed method against recent video matting methods~\cite{sharianvideo,johnsontip,bai2011towards,bai2009video} on \emph{Amira}, \emph{Face} and \emph{Woman} sequences from the dataset. Additional comparisons and video are available at the url\protect\footnotemark[1].\footnotetext[1]{\url{https://goo.gl/Ho5xMN}} The yellow arrows indicate the regions of discontinuity between consecutive frames. The low contrast between the foreground and the background in \emph{Amira} is a challenging scenario for most matting algorithms. Laplacian-based smoothing used by existing methods produce ambiguity near the boundaries as pixel neighbors tend to be unreliable for accurate propagation. The use of patch-based neighbors in the proposed method enables us to remove such artifacts in the final matte. The sequences in Fig.~\ref{fig:Qualitative2} represent cluttered background which is challenging for most sampling-based algorithms. Our error based matte is accurately able to distinguish between the hair and the background when compared to the ground truth, showing its effectiveness in highly textured regions.
\begin{table}[t]
\centering
\caption{Comparison of temporal jitter error rates of different video matting algorithms against the proposed method}
\label{tab:quantitative}
\resizebox{0.5\textwidth}{!}{%
\begin{tabular}{cccccc}
\hline
Video  & SC~\cite{bai2009video} & BA~\cite{bai2011towards} & EH~\cite{sharianvideo} & JO~\cite{johnsontip} & Proposed \\ \hline
\textit{Face}      & 3.46             & 2.92                      & 4.26                         & 2.37                           & \textbf{1.49}            \\ 
\textit{Dancer}    & 4.72             & 4.13                      & 1.48                         & 2.13                           & \textbf{1.46}            \\ 
\textit{Arm}       & 4.43             & 2.91                      & 2.54                         & 3.52                           & \textbf{1.58}            \\ 
\textit{Woman}     & 4.03             & 2.72                      & 3.36                         & 2.82                           & \textbf{2.05}            \\ 
\textit{Smoke}     & 3.17             & 2.96                      & \textbf{1.80}                & 4.85                           & 2.19                     \\ 
\textit{Cat}       & 2.54             & 4.18                      & 2.45                         & 4.41                           & \textbf{1.40}            \\ 
\textit{Chimp}     & 3.54             & 4.63                      & 2.90                         & 2.09                           & \textbf{1.81}            \\ 
\textit{Girl}      & 4.55             & 4.34                      & 2.31                         & 2.12                           & \textbf{1.65}            \\ 
\textit{Whitegoat} & 3.72             & 3.85                      & 3.47                         & 2.17                           & \textbf{1.76}            \\ 
\textit{Amira}     & 4.18             & 4.27                      & 2.72                & 2.09                           & \textbf{1.72}                     \\ 
\textit{Girl2}     & 4.54             & 4.40                      & 2.18                          & 2.0                 & \textbf{1.86}                     \\
\textit{Office} & 3.94             & 3.64                      & 2.94                         & \textbf{2.05}                           & 2.41            \\ 
 
\textit{Soccer}    & 4.05             & 3.24                      & \textbf{2.31}                & 2.59                           & 2.79                     \\ 
\textit{Unicorn}   & 3.23             & 3.21                      & 2.91                         & 3.34                           & \textbf{2.28}            \\ 
\textit{Dog}       & 4.09             & 3.62                       & 3.5                & 2.04                           & \textbf{1.73}                     \\ \hline
\end{tabular}%
}
\end{table}
\subsection{Quantitative evaluation} 
We perform quantitative comparison to evaluate the temporal coherence of the extracted mattes by measuring the difference in alpha values between successive frames as in \cite{lee2010temporally}. The temporal flicker at the $i^{th}$ pixel in frame $t$ is measured as
\begin{equation}
f_i(t) = \frac{\left |\alpha_i(t+1)-\alpha_i(t) \right |}{\left |I_i(t+1)-I_i(t)\right |},
\end{equation}
where $\alpha_i(t)$ and $I_i(t)$ are the alpha and RGB color values at pixel $i$ in frame $t$.


Table~\ref{tab:quantitative} compares the mean temporal jitter error across 15 sequences in the dataset with recent video matting approaches. As can be seen, the proposed method is able to produce the least temporal jitter across most of the sequences. For the few exceptions, the reconstruction error cannot be trustworthy when the true $F$ and $B$ samples are not present in the dictionary leading to poor initial estimates. Apart from the smooth reconstruction error formulation, the use of a patch-based coherency sensitive hashing is instrumental in the increased performance of the proposed method. \cite{johnsontip} uses pixel neighbors in its graph formulation which can be erroneous due to noise. Moreover, the smoothness of alpha is not maintained in the feature vector for sparse coding in \cite{johnsontip}.

\textit{Runtime performance}: 
Table~\ref{tab:runtime} compares the running time of the proposed method with other sampling-based video matting approaches. MATLAB implementations  were evaluated on a PC running Intel Xeon 3.2 GHz processor. The proposed method perform comparable to the current approaches without compromising on the quality of the matte.                                   	  

\begin{table}[t]
\centering
\caption{Comparison of running time of the proposed method with recent sampling-based approaches}
\label{tab:runtime}
\resizebox{0.5\textwidth}{!}{%
\begin{tabular}{cccccc}
\multicolumn{1}{l}{} & \multicolumn{1}{l}{} & \multicolumn{1}{l}{}                                    & \multicolumn{3}{c}{Total Time (secs)}      \\ \hline
Video                & Size                 & \begin{tabular}[c]{@{}c@{}}No. of\\ Frames\end{tabular} & EH {[}7{]} & JO {[}10{]}   & Proposed      \\ \hline
Smoke                & 500x500              & 90                                                      & 4491       & 3798          & \textbf{3628} \\
Arm                  & 640x540              & 49                                                      & 3260       & \textbf{1618} & 2188          \\
Dancer               & 480x360              & 40                                                      & 5488       & 2803          & \textbf{2589} \\
Face                 & 640x540              & 78                                                      & 5378       & 4955          & \textbf{4786} \\
Archaeology          & 480x405              & 128                                                     & 4980       & \textbf{2524} & 2961          \\
Woman                & 450x400              & 154                                                     & 5541       & \textbf{2912} & 3178          \\ \hline
\end{tabular}
}
\end{table}

\section{Conclusion}
We present a novel video matting framework that treats the matting problem from the perspective of reconstruction error of a feature. Foreground and background dictionaries, whose bases are used to reconstruct an unknown feature vector with L1-regularization are used to measure the error towards $F$ and $B$ respectively. A NLM framework is also proposed that is integrated across multiple frames to ensure temporal coherence in the video mattes. Experimental evaluations demonstrate that the proposed method has advantages over current matting methods that use a Laplacian based smoothing.


%





\ifCLASSOPTIONcaptionsoff
  \newpage
\fi



\bibliographystyle{IEEEtran}
\bibliography{IEEEabrv,IEEEexample}
\end{document}